\documentclass[journal,comsoc]{IEEEtran}
\usepackage{amssymb,amsmath}
\usepackage{cite}
\usepackage{graphicx}
\usepackage{psfrag}
\usepackage{mathrsfs} 
\usepackage{tabularx}
\usepackage{multirow}
\usepackage{graphicx}
\usepackage{url}
\usepackage[latin1]{inputenc}
\usepackage[absolute,overlay]{textpos}
\usepackage{gensymb}
\usepackage{cases}
\usepackage{graphicx}
\usepackage{float}
\usepackage[linesnumbered,ruled,lined]{algorithm2e}
\renewcommand{\arraystretch}{0.7}
\usepackage{pgf}
\renewcommand{\arraystretch}{1.25} 

\usepackage[nocomma]{optidef}

\usepackage{amsthm}

\newtheorem{remark}{Remark}

\usepackage{booktabs}
\usepackage{color,soul}

\begin{document}

\title{AUV Trajectory Learning for Underwater Acoustic Energy Transfer and Age Minimization}
\author{
	\IEEEauthorblockN{Mohamed Afouene Melki, Mohammad Shehab, and Mohamed-Slim Alouini \\
	}
	\thanks{This work is supported by the KAUST Office of Sponsored Research under Award ORA-CRG2021-4695.}
	\thanks{The authors are with CEMSE Division, King Abdullah University of Science and Technology (KAUST), Thuwal 23955-6900, Saudi Arabia (emails: mohamed.melki@kaust.edu.sa, mohammad.shehab@kaust.edu.sa, slim.alouini@kaust.edu.sa).}
}
\maketitle


\begin{abstract}

Internet of underwater things (IoUT) is increasingly gathering attention with the aim of monitoring sea life and deep ocean environment, underwater surveillance as well as maintenance of underwater installments. However, conventional IoUT devices, reliant on battery power, face limitations in lifespan and pose environmental hazards upon disposal. This paper introduces a sustainable approach for simultaneous information uplink from the IoUT devices and acoustic energy transfer (AET) to the devices via an autonomous underwater vehicle (AUV), potentially enabling them to operate indefinitely. To tackle the time-sensitivity, we adopt age of information (AoI), and Jain's fairness index.  We develop two deep-reinforcement learning (DRL) algorithms, offering a high-complexity, high-performance frequency division duplex (FDD) solution and a low-complexity, medium-performance time division duplex (TDD) approach. The results elucidate that the proposed FDD and TDD solutions significantly reduce the average AoI and boost the harvested energy as well as data collection fairness compared to baseline approaches. 

\end{abstract}

\section{Introduction}
The health of our planet relies heavily on the well-being of its oceans, which serve as vital sources of oxygen, sustenance, and energy. Communication in extreme environments such as seas and deep oceans is of paramount importance due to its role in search and rescue operations, surveillance, maintenance of optical fibers and gas pipelines, and environmental monitoring among many other applications \cite{centelles2015wireless, extreme}. Each application has specific requirements in terms of communication range, energy consumption, data rate, and latency tolerance. Over time, underwater connectivity is becoming more vital.


The Internet-of-Underwater-Things (IoUTs), a global network connecting underwater objects, facilitates continuous monitoring of oceans. This network generates high-resolution data essential for training machine learning (ML) algorithms to swiftly evaluate potential climate change solutions and aid in decision-making processes. In addition, recent advances highlighted the potential of IoUT for underwater localization \cite{loc2} and the defense against attacks for such localization approaches \cite{article}. Furthermore, and within the efforts leading to enhanced underwater connectivity for divers, the authors of \cite{shihada2020aqua} developed the world's first underwater wifi "Aqua-Fi" to transfer data from a diver's smartphone to a gateway device mounted on their equipment. This gateway then relays the data through a light beam to a surface computer, which is connected to the internet via satellite. However, the light beam is only suitable for short distances and suffers from disruptions due to tides and underwater movements. In response to these limitations, recent advancements have explored alternative acoustic approaches to achieve long-range, low-power underwater communication. For instance, the work in \cite{eid2023enabling} presents an underwater backscatter communication system that enables long-distance communication by encoding data in sound waves reflected back to a receiver. Tested in river and ocean environments, this system demonstrated a significantly greater communication range than previous technologies, while maintaining low power consumption, marking a substantial improvement for sustained underwater connectivity.


Due to their ability to navigate freely and reach remote and risky locations for divers, AUVs are considered to be a very attractive solution for gathering information from IoUT devices. In this context, the work in \cite{Safeer2024} illustrated the application of federated learning in collaborative information processing via AUVs in IoUT. Nevertheless, the nature of underwater communication is different from other environments. For instance, conventional IoUT sensors, reliant on battery power, face limitations in lifespan, difficulties with battery replacement, and pose environmental hazards upon disposal. Unlike terrestrial systems, underwater communication suffers from significant signal attenuation, tidal turbulence and multi-path effects. These challenges necessitate innovative approaches to optimize the paths of AUVs for effective data collection and wireless energy transfer.

Understanding and overcoming the limitations of communication in underwater environments can lead to more robust and efficient systems, paving the way for future innovation in the field. In this context, the work in \cite{pal2022communication,Survey} presented a comprehensive summary on different communication and energy transfer technologies underwater. However, no single technology can meet all requirements across different activities simultaneously. Hence, each system is tailored to suit specific applications. According to their discussion and many other works such as \cite{9217956}, it is evident that despite the drawback of low data rates and slow propagation, the adoption of acoustic transmission is superior to optical transmission and magnetic induction in terms of resilience and long range. This is further elaborated in \cite{9217956}, where the attenuation coefficients for magnetic induction and electromagnetic-based energy transfer are too high compared to AET. This in turn affects the efficiency of energy transmission causing a great loss when the distance of energy transfer is longer than few meters. This preference is further supported in \cite{zia2021state}, which provides a comparative analysis of both commercial and research acoustic modems. This study highlights the modem parameters crucial for underwater applications, such as operating range, data rate, modulation schemes, and power consumption, identifying current trends and key design challenges. Additionally, the survey in \cite{Jouhari} emphasizes the importance of acoustic technology for underwater communication systems, particularly as underwater IoUT devices typically transmit small, delay-tolerant data packets such as sensor readings, making acoustic technology well-suited for our endeavor. Despite the infancy of the idea of underwater AET, an example of new advances in hardware design for AET can be found in \cite{ultrasonic}, which illustrates transducer structure for underwater AET system.


Assuming instant reception, AoI is defined as the time elapsed since the reception of the last information update from an IoUT device \cite{AoI,AoI_plus}. AoI minimization is of paramount importance since the lower the AoI, the fresher the information received from a specific IoUT device. A lot of research has been conducted on applying machine learning in order to minimize the AoI of devices in above-water environments. For instance, the authors of \cite{Magid} applied deep reinforcement learning (DRL) to design the trajectory of UAV collecting information from a small number of grounded devices with the goal of minimizing the AoI of these devices. Moreover, the study in \cite{9950310} explored the use of multiple Unmanned Aerial Vehicles (UAVs) to optimize both AoI and power consumption of a relatively larger number of devices. In \cite{swarm}, the same authors expanded their setup further, where a swarm of UAVs collects data from device clusters constituting a massive IoT network. These efforts highlight the importance of efficient path planning, scheduling, and power management to ensure timely data collection and energy efficiency. 

To this end, authors generally resorted to RL due to the high complexity and dimensionality of the trajectory optimization problem in such cases, especially when taking the energy and environmental details into consideration. In particular, when dealing with problems of high dimensional state and action spaces, DRL is known for its efficiency. This is because DRL is able to sample and reduce the dimensionality of those spaces, which renders a faster, yet highly sub-optimal solution. Similar research in underwater environments remains sparse. This area has not been extensively studied until Omoke et al. \cite{10268591} published the first attempt to apply RL to perform simultaneous wireless power transfer and data collection in an underwater system. They demonstrated relatively good results, despite their setup being somewhat similar to terrestrial systems, where a vehicle needs to get very close to devices to power them up and exchange data. However, their model did not account for AoI and its trade-off with the energy harvesting goal.

Hence, the exploration of RL in underwater environments is crucial due to the unique challenges posed by underwater communication and navigation. In this context, the authors of \cite{10012479} presented a wide discussion on how RL could help tackle IoUT challenges. The online nature of RL was shown to be well-tailored to the dynamic nature of underwater channel, link outages, bandwidth allocation as well as AUV control problems. Moreover, in \cite{Relay}, Dai et al. proposed an RL approach for underwater relay selection and power control. By focusing on RL to optimize AUV trajectories, this work aims to address the challenge of battery-constrained IoUT and information uplink, which contributes to the advancement of underwater wireless communication technologies. We assume a set of underwater IoT devices that rely mainly on harvested energy from acoustic sources to transmit information The target is to optimize the underwater AUV trajectory and device scheduling in order to minimize the average AoI and implicitly maximize the harvested energy via AET. 
\subsection{Contributions}

The motivation behind this work stems from its novelty in accounting for AoI minimization and solving the problem of battery replacement for IoUT devices. This leads to avoiding hazardous material underwater, which might affect sea life. Our research also complements the acoustic communication system in a sustainable framework. Particularly, in the context of planning paths and scheduling for underwater simultaneous wireless communication and energy transfer, the contributions of this paper are summarized as follows:

\begin{itemize}
    \item With the goal of minimizing the average weighted AoI of IoUT devices, we develop a couple of DRL solutions for AUV trajectory planning for simultaneous AET and information uplink  
    \item The first approach is FDD-based high performance, high complexity, and two antenna-based solution.
    \item The second one is a TDD-based low complexity, low-cost alternative solution that adopts only one antenna for both AET and information reception and the AUV.
    \item Compared to baseline approaches such as random walk (RW), round robin (RR), and the greedy algorithm (GA), the proposed FDD and TDD DRL-based approaches jointly minimize the AoI and maximize the harvested energy and data collection fairness at the IoUT devices.
\end{itemize}

\subsection{Outline}
The rest of the paper is organized as follows: Section~\ref{sec:system} describes the system model and basic information about underwater communication. Next, Section~\ref{sec:problem} formulates the simultaneous AET and information uplink problem and elucidates TDD and FDD. Section~\ref{sec:solution}  proposes the DRL-PPO, while Section~\ref{sec:results} depicts the experimental results. Finally, Section~\ref{sec:conclusions} concludes the paper. The appendices include some mathematical proofs and derivations that serve the solution of the presented optimization problem.

\section{System Model}\label{sec:system}
\begin{figure}[t!]
	\centering
	\includegraphics[width=0.95\columnwidth]{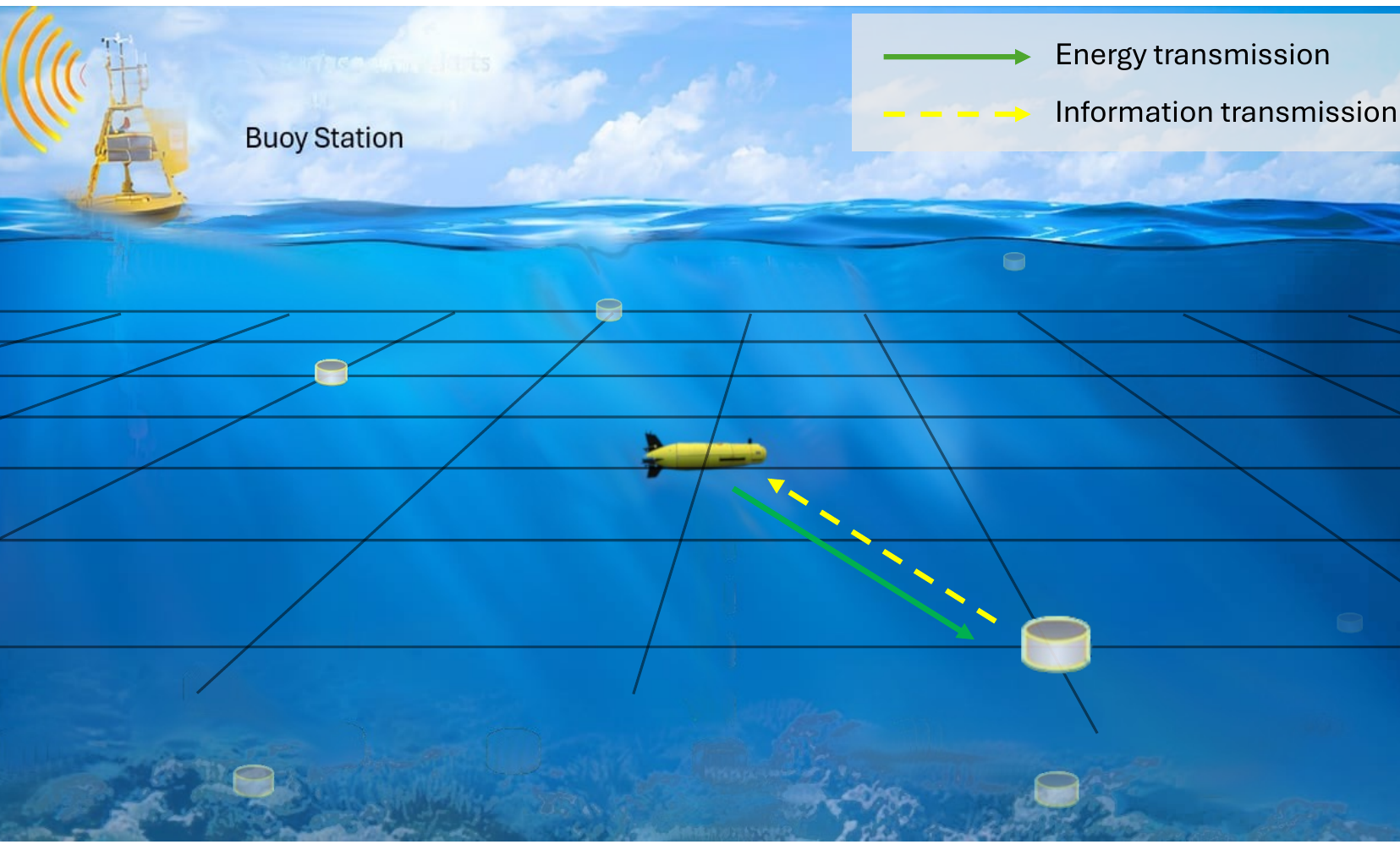}
	\caption{AUV and IoUT network in the underwater grid world} 
	\label{system model}
\end{figure}

\subsection{Network Model}


As illustrated in Fig. \ref{system model}, we consider an IoUT network, comprising an AUV deployed from a surface buoy station, and $K$ IoUT nodes dispersed randomly in a 3-D underwater space. Each IoUT device $k$ is spatially distributed across the 3D equally spaced grid world and is assigned fixed coordinates $c_k=(x_k,y_k,z_k)$. Similarly, at time $t$ the AUV is designated coordinates $l_{auv}(t)=(x_{auv}(t),y_{auv}(t),z_{auv}(t))$. Each IoUT device is equipped with sensors designed for monitoring essential underwater parameters such as temperature, pH level, and dissolved oxygen concentration. The AUV maneuvers through the grid world to establish communication with the underwater sensor nodes, where communication is facilitated through an acoustic modem/hydrophone.
Moreover, as the AUV traverses the network, it employs AET to provide energy, thereby recharging energy-constrained IoUT nodes.

\subsection{Channel Model}
In acoustic communication, the received level (RL) in dB at an IoUT device that is located at distance $d$  from the source (i.e, AUV) is calculated as \cite{6266681}
\begin{equation}
 \label{RL}
RL=SL-AL-NL,
\end{equation}
where $SL$ is the acoustic source level, $AL$ characterizes the total attenuation level and $NL$ is the ambient noise level. The SL of an underwater acoustic transmitter is given by 
\begin{equation}
\label{SL}
SL=170.8+10\log_{10}P_{elec}+10\log_{10}\eta+DI,
\end{equation}
where DI is the directivity index of the source in dB,  $P_{elec}$ is the electrical input power at the source, while the electro-acoustic power conversion efficiency $\eta$ varies between 0.2 and 0.7 for practical modems.

Assuming deep water characteristics and neglecting reflection from the air and bottom surfaces throughout the analysis, combining absorption, channel spreading loss, and noise, the total attenuation level (AL) in dB is given by\footnote{For simplicity and focusing on the main system optimization, we assumed flat fading}
 \begin{equation}
 \label{AL}
AL=k_{s}\cdot10\log_{10}d+d\cdot10\log_{10}\alpha\left(f\right),
\end{equation}
where and $k_s$  is the spreading factor that takes the value of 1 or 2 depending on the assumptions; $k_s = 2$ is referred to as spherical spreading experienced when a sound wave propagates away from the source uniformly in 3 directions, when $d < \text{water depth}$. Meanwhile, $k_s = 1$ is referred to as cylindrical spreading experienced when the acoustic signal systematically hits the sea surface and sea floor before reaching the destination.  $\alpha\left(f\right)$ is the absorption coefficient of acoustic waves underwater that can be expressed using Thorp's formula 
for frequencies above a few hundred Hz, \cite{9217956}
\begin{equation}
\label{alpha(f)}
 \alpha\!\left(f\right)\!=\! 0.11 \! \frac{f^2}{f^2+1} \!+ 44  \frac{f^2}{f^2+4100} \!+ 2.75 \times 10^{-4}f^2 \!+\! 0.003,
\end{equation}
where $f$ is the acoustic transmission frequency.

\section{Problem formulation}\label{sec:problem}
\subsection{Acoustic Energy Transfer and Information Uplink}

\subsubsection{Acoustic Energy Transfer}
The process of AET begins with the emission of acoustic waves by the AUV through its hydrophone. These waves carry transmitted power and propagate as fluctuating pressure waves characterized by their amplitude, frequency, and phase. Subsequently, the IoUT nodes' hydrophones receive these energy waves. The process also involves the use of transducers that can convert acoustic pressure waves into electrical energy. This electrical energy is then stored in a battery onboard the IoUT node. The pressure fluctuations can be expressed as \cite{6266681} 
 \begin{equation}
 \label{pressure p}
     p=10^{{RL}/{20}}.
 \end{equation}
The fluctuations generate a voltage at its open circuit terminals as mentioned earlier. The receiving voltage sensitivity (RVS) of a hydrophone is defined as \cite{6266681} 
 \begin{equation}
 \label{rvs}
     RVS=20\log_{10}M,
 \end{equation}
where $M$ is the sensitivity in $V / \mu Pa$. Using \eqref{pressure p} and \eqref{rvs}, the induced voltage, at the receiver hydrophone terminals $ V_{ind}$ is given by 
\begin{equation}
\label{v ind}
V_{ind}=p\cdot M =10^{\frac{RL+RVS}{20}}.
\end{equation}

The electrical power available for harvesting, $P_{available} $ depends on
the impedance matching between the receiver hydrophone and
the surrounding seawater. For a single hydrophone, this can be
expressed as
\begin{equation}
\label{p available}
    P_{available}={{V_{ind}^{2}}\over{4R_{p}}},
\end{equation}
where $R_{p}$ is the load resistance required to ensure impedance matching. This gives us the total harvestable power, which is given by
\begin{equation}
\label{p harvested}
    P_{harv}=\eta \cdot P_{available}=\eta \cdot{{10^{{\left({RL+RVS}\right)}/{10}}}\over{4R_{p}}},
\end{equation}
where $\eta$ is the acoustic-electric power conversion efficiency. Then, the energy harvested is calculated as
\begin{equation}
\label{E harvested}
     E_{harv}= P_{harv}\cdot  \tau_{charging},
\end{equation}
where $ \tau_{charging}$ represents the time taken by the AUV to transmit energy to the IoUT node.

\subsubsection{Information Uplink}
During the communication between the IoUT nodes and the AUV, the latter collects data. The SNR, $\gamma_{k}$ between the $k^{th}$ node and the AUV can be expressed as
\begin{equation}
\label{snr}
    \gamma_{\mathrm{k}} = 2^{\left({{\mathcal{S}}}/{B}\right)} - 1,
\end{equation}
where ${\mathcal {S}}$ is the system throughput and $B$ is the bandwidth.  The transmit power of the $k^{th}$ node $P_{trans,k}$ is \cite{10268591} 
\begin{equation}
\label{p transmit}
   P_{trans,k} =\gamma _{k} \cdot 10^{\frac{NL}{10}} \cdot 10^{\frac{AL_{k,auv}}{10}},
\end{equation}
where $AL_{k,auv}$ is the transmission loss between the $k^{th}$ node and the AUV. The energy required by a node $k$ to transmit information to the AUV is 
\begin{equation}
\label{E transmited}
         E_{req,k}= P_{trans,k}\cdot  \tau_{data},
\end{equation}
where $\tau_{Data}$ is the time required to transmit the payload from the sensor to the AUV.

\subsubsection{Age Of Information}
Upon transmission of a packet by a selected IoUT device 
$k$ at time step $t$, its AoI $A_k(t)$ resets to 1, which indicates that fresh information has just been received from that device. 
\subsection{Duplexing Techniques} 
\begin{figure}[t!]
	\centering
	\includegraphics[width=0.98\columnwidth]{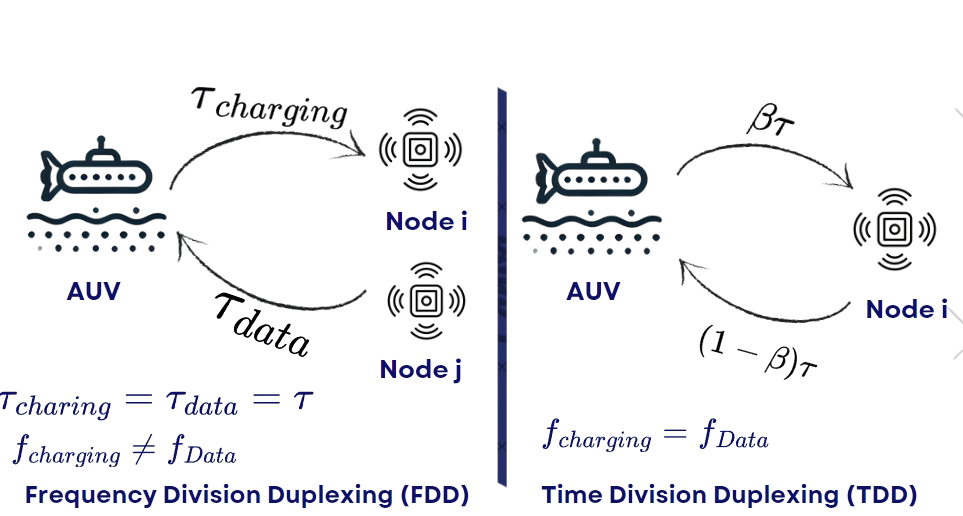}
	\caption{AET and information uplink using FDD vs TDD } \vspace{0mm}
	\label{FDD TDD}
\end{figure}

In underwater simultaneous communication and AET, 
duplexing techniques play a crucial role in determining the efficiency and effectiveness of data and energy transmission. Herein, we adopt two primary types of duplexing, namely FDD and TDD. Each technique has its own set of advantages and challenges as noted in \cite{ahmadian}, which impacts their applicability in various circumstances. The two techniques are illustrated in Fig. \ref{FDD TDD}
\subsubsection{Frequency-Division Duplexing}
    FDD is a duplexing method where separate frequency bands are allocated for AET and information uplink. This means that AET and data transmission can occur simultaneously but on different frequencies. FDD is particularly useful in scenarios where separate and large uplink and downlink bandwidth requirements are needed, providing continuity and high capacity for energy and information transmission with limited interference.
    As shown in the left part of Fig. \ref{FDD TDD}, the AUV utilizes two distinct frequency bands with different central frequencies $f_{charging}$ for AET to $i^{th}$ node and $f_{data}$ for data uplink from $j^{th}$ node.This separation of frequencies ensures that the two processes can occur without interference.
    In our case, FDD can be advantageous due to its ability to handle simultaneous bidirectional AET and information uplink, reducing latency and improving throughput. However, the need for separate frequency bands can lead to spectrum inefficiency, as it requires a broader range of frequencies to operate. Additionally, the hardware complexity increases because of the necessity for duplexers as well as multiple antennas to separate the data and AET at the transmitter and receiver. 
\subsubsection{Time-Division Duplexing}
TDD is a method where the same frequency band is used for both AET and information uplink but at different times. This technique involves switching between transmitting and receiving modes within designated time slots, effectively utilizing the same frequency spectrum for both operations. TDD is particularly useful in scenarios where limited hardware and frequency resources exist, allowing for dynamic adjustment of AET and information uplink based on real-time demand.
As shown in the right part of Fig. \ref{FDD TDD}, the AUV shares the same frequency band for AET and data uplink but allocates different durations for each operation. The total communication duration 
$\tau$ is divided into  $\beta\tau$ for AET and $(1-\beta)\tau$ for data uplink, where $\beta$ represents the time-splitting factor. This allocation ensures efficient use of the available spectrum by sequentially performing energy transfer and data communication.
TDD is advantageous in the sense of its efficient spectrum usage and flexibility. It simplifies the hardware design by eliminating the need for separate frequency bands, antennas, and duplexers. However, precise time synchronization is crucial to avoid interference between data uplink and energy downlink transmissions.

\subsection{Problem Formulation}\label{INF}
Inspired by existing works such as \cite{Magid}, \cite{swarm}, and \cite{10032494} which focus on minimizing the AoI and improving fairness in UAV-based communication systems, we extend these concepts to address the specific challenges in AUVs.To minimize the weighted average AoI of collected data, the AUV trajectory planning problem can be formulated as follows\footnote{In this problem formulation, we focus on the communication between the AUV and sensor nodes. The dynamics of the AUV and its energy consumption are beyond the scope of this work. For simplicity, we assume the AUV has sufficient energy throughout the entire navigation time.} 
\begin{subequations}\label{P1}
\begin{alignat}{2}
\mathbf{P1:} \quad 
&\underset{\{x_k(t)\}_{t=0}^{T},\, \{\boldsymbol{l}_{auv}(t)\}_{t=0}^{T}}{\min}  
&& \frac{(1-\mathcal{J}_T)}{T} 
\cdot \sum_{t=1}^{T} \frac{1}{K} \sum_{k=1}^{K} 
\delta_k \cdot A_k(t) \notag \\
&\text{Subject to:} \quad 
&& A_k(t) \leqslant A_{\text{max}}, 
\quad \forall k,\; \forall t \notag \\
&&& \mathcal{J}_T \geqslant \mathcal{J}_{T_{\min}} \notag \\
&&& x_k(t)\, e_k(t) \geqslant x_k(t)\, E_{\text{req},k}(t),  \forall k,\; \forall t \notag \\
&&& \sum_{k=1}^{K} x_k(t) = 1, 
\quad \forall t  \notag \\
&&& x_k(t) \in \{0,1\}, 
\quad \forall k,\; \forall t  \notag \\
&&& \boldsymbol{l}_{auv}(t) \in \mathcal{X}, 
\quad \forall t  \notag
\end{alignat}
\end{subequations}

In this formulation, $\delta_k$ represents the importance weight assigned to device $k$ and $A_{\text{max}}$ is the maximum allowed AoI in the system. The binary variable $x_k$ indicates whether device $k$ is selected for data transmission ($x_k = 1$) or not ($x_k = 0$). The energy stored in device $k$ at time $t$ is denoted by $e_k(t)$, while $E_{\text{req},k}(t)$ represents the minimum energy required for device $k$ to transmit data. The constraint $x_k \cdot e_k(t) \geqslant x_k \cdot E_{\text{req},k}(t)$ ensures that the selected device has enough energy to transmit. The condition $\sum_{k=1}^{K} x_k = 1$ enforces that only one device is selected at each time step. Finally, the constraint $l_{auv}(t) \in \mathcal{X}$ ensures that the AUV remains within the defined grid world, where $\mathcal{X}$ denotes the set of all possible locations within the grid.

The weighted average AoI, as expressed in the objective function, allows assigning varying levels of importance to different nodes. This is particularly useful in scenarios involving critical and non-critical nodes, where higher priority can be given to nodes with stricter data freshness requirements \cite{9692982}. In this work, we simplify the analysis by setting all weights $\delta_k$ to 1, ensuring that all nodes are treated equally in terms of importance. This assumption allows us to focus on evaluating the overall performance of the proposed approach without introducing additional bias in the optimization objective.

The main objective of problem \textbf{P1} is to minimize the average AoI. To achieve this, the solution should ensure that sufficient energy is available at each IoUT node for data transmission, which implicitly maximizes the harvested energy. However, this approach may lead to an unfair distribution of data collection among devices despite boosting the overall global performance. To address this issue, we make use of Jain's fairness index, $\mathcal{J}_T$, as a metric to ensure fair data gathering \cite{10032494}. This index is defined based on the frequency of data collection from each device over a period $T$, where 
\begin{equation}
\label{jain fairness J}
\mathcal{J}_T = \frac{\left(\sum_{k=1}^{K} D_T(k)\right)^2}{K \sum_{k=1}^{K} D_T(k)^2},
\end{equation}
where $D_T(k)$ denotes the total number of times the data was collected from the device $k$ over the period $T$. 
The value of $\mathcal{J}_T$ lies within the range [0, 1], and it can be interpreted as follows:  
\begin{itemize}
    \item If $\mathcal{J}_T$ is close to 1, it indicates that all nodes are receiving equal attention during data collection, meaning the data collection process is balanced and fair.  
    \item If $\mathcal{J}_T$ is close to 0, it implies an unbalanced distribution where some nodes are prioritized significantly more than others, leading to neglect of certain nodes.  
\end{itemize}
In the context of this work, Jain's fairness index ensures that no IoUT node is overly prioritized or ignored, even as the global performance (e.g., minimizing AoI) is optimized. By monitoring $\mathcal{J}_T$, we achieve a balance between performance and fairness, which is crucial in applications where uniform data collection across nodes is essential,

In our problem, we aim to maximize Jain's fairness index. Moreover, this index is not allowed to fall below a specific threshold $\mathcal{J}_{{min}}$ to guarantee a reasonable amount of fairness on data collection. 

In summary, our objective is to simultaneously achieve fair data gathering and minimize the average AoI. Given the complexity and NP-hard nature of this problem, traditional optimization techniques are impractical. To model the association and interaction pattern between the AUV and IoUT devices, the AUV trajectory planning problem is formulated as a Markov decision process (MDP) that captures the dynamics of the AUV. DRL is employed for this MDP due to its ability to manage large state and action spaces and adapt to complex, nonlinear systems, robustly optimizing the AUV's trajectory while ensuring energy-efficient communication and fair data collection across the network.

\section{The proposed DRL solution}\label{sec:solution}

\subsection{Markov Decision Process Formulation}

\begin{figure}[t!]
	\centering
	\includegraphics[width=0.90\columnwidth]{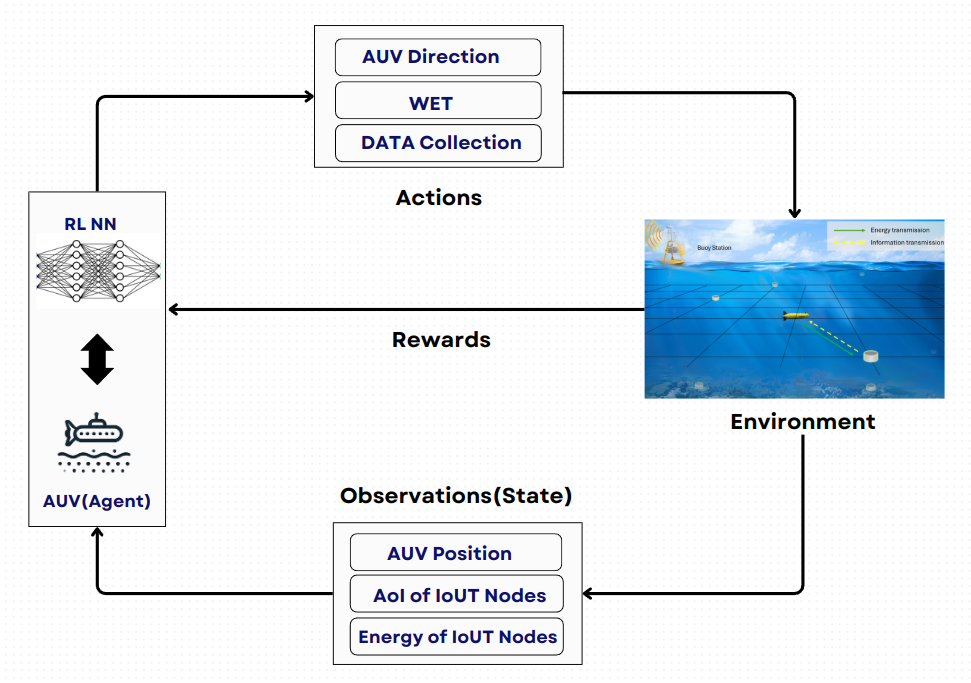}
	\caption{The interaction between the agent and the environment } \vspace{0mm}
	\label{Interaction Agent ENV}
\end{figure}

We formulate the problem as an MDP that is defined by the tuple $\langle \mathcal{S}, \mathcal{A}, R, P \rangle$, where $\mathcal{S}$ is the state space, $\mathcal{A}$ represents the action space, $R$ denotes the reward function. We consider a finite horizon MDP with a probabilistic state transition function $P: \mathcal{S} \times \mathcal{A} \times \mathcal{S} \rightarrow \mathbb{R}$. At time instant $t$, the agent (AUV) observes the current state $s(t)$ from the environment and tries to follow the optimal policy by selecting the best action $a(t)$, which maximizes the reward $r(t)$ and transiting to the next state $s(t+1)$ with a probability $p(s(t),s(t+1))$. The agent can then rely on these policies to make all future data transfer and AET decisions as illustrated in Fig. \ref{Interaction Agent ENV}. For convenience, we propose an episodic MDP. This means that the AUV's data collection and energy transmission is over at time $T \in \mathbb{N}$, where the total navigation time $T$ is dictated by the amount of navigation energy available at the AUV and the dimensions of the network. Time slots are discretely divided as $t=\{\tau, 2\tau, \dots, T\}$, where $\tau$ is the time that the AUV takes to move from one grid point to another adjacent one. 
$\tau$ is dictated by the AUV's velocity and the spacing between grids.
\subsubsection{State space}
The state space of the system at time slot $t$ is defined as $s(t) = (\boldsymbol{l_{auv}}(t),\boldsymbol{A}(t), \boldsymbol{\mathcal{E}}(t) )
$, where $\boldsymbol{l_{auv}}(t)$  is a vector containing the current position of the AUV at time slot $t$. $\boldsymbol{A}(t) $ is a vector that contains the current AoI of all the IoUT devices at time slot $t$, where $A_k(t)\in\mathcal{I} = [1,2,..., A_{max}]$ and $A_{max}$ is the maximum allowed AoI in the model, which is chosen to be arbitrarily high. $\boldsymbol{\mathcal{E}}(t)$ represents a vector of the available energy at each IoUT node, which belongs to a discrete set $\mathcal{E}$. . Without loss of generality, we will consider the discrete 3D coordinates of the AUV and IoUT to be integers to reduce the computational demand of representing the system state. Note that the state is updated before being fed to the agent. Considering the three described components, the dimensionality of the state space could be described as
\begin{equation}  
\label{state space}
\mathcal{S}=\underbrace{\mathcal{X}^{3}}_{\text{ Auv Position}}\times \underbrace{\mathcal{E}^{K}}_{\text{Energy Stored}}\times \underbrace{\mathcal{I}^{K}}_{\text{Device age of information}}, \end{equation}
\begin{remark}
These components were selected to capture the critical spatial (\(\boldsymbol{l_{auv}}(t)\)), temporal (\(\boldsymbol{A}(t)\)), and energy (\(\boldsymbol{\mathcal{E}}(t)\)) dynamics of the system, enabling the agent to make informed decisions for AoI minimization and fairness. Excluding \( \boldsymbol{\mathcal{E}}(t) \) from the state space results in failed data transmissions, as the agent is unable to account for energy availability at the nodes. Similarly, removing \( \boldsymbol{A}(t) \) leads to poor AoI optimization, as the agent lacks awareness of data freshness.
\end{remark}
\subsubsection{Action space}   We have developed two approaches for defining the action space in our RL framework: a 3D action space for FDD communication and a 2D action space for TDD communication. \\
\textbf{First Approach: 3D Action Space (FDD)} 

The AUV action at time slot \( t \) is defined as \( a(t) = (d(t), W(t), I(t)) \), where \( d(t) \) represents the movement of the AUV in a given direction, \( W(t) = (w_1(t), w_2(t), \ldots, w_K(t)) \) is a sparse vector representing the node chosen for WET with \( w_k(t) = 1 \) indicating that node \( k \) is selected, and \( I(t) = (i_1(t), i_2(t), \ldots, i_K(t)) \) is a sparse vector representing the node chosen for Information transmission with \( i_k(t) = 1 \) indicating that node \( k \) is selected for information uplink. The action space is given by:
\begin{equation} 
\label{action space fdd}
\mathcal{A} = \underbrace{\mathbb{D}^1}_{\text{direction}} \times \underbrace{\{0,1\}^K}_{\text{sparse vector for WET}} \times \underbrace{\{0,1\}^K}_{\text{sparse vector for data collection}},
\end{equation}
where $\mathbb{D}$ is the direction of movement that can be right, left, up, down, forward, or backward. In this approach, the AUV uses different bands (i.e., central frequencies) for AET and information uplink via two separate antennas, which allows simultaneous operation with different devices.\\   
\textbf{Second Approach: 2D Action Space (TDD)} 

To simplify decision-making and reduce computational and hardware demands, TDD reduces the action space to two decisions. The AUV selects a direction for movement and selects the same node for both AET and information uplink, with a portion of the time $\beta \tau$ dedicated to charging and $(1-\beta)\tau$ to data collection. Thus, the TDD action space is defined as: 
\begin{equation} 
\label{acion space tdd}
\mathcal{A} = \underbrace{\mathbb{D}^1}_{\text{direction}} \times \underbrace{\{0,1\}^K}_{\text{sparse vector for node selection}},
\end{equation}
\begin{remark}
The action space components were chosen to allow the agent to balance AoI reduction, energy replenishment, and efficient navigation. By independently selecting nodes for WET (\(W(t)\)) and data collection (\(I(t)\)) in the FDD approach, or combining them into a single node decision in the TDD approach, the framework enables efficient decision-making while accounting for the trade-offs between performance and complexity. Reducing the dimensionality of the action space in the FDD approach simplifies decision-making but compromises performance, as the agent loses the ability to independently optimize WET and data collection. 
\end{remark}
where the sparse vector indicates the IoUT device selected for both AET and information, with the \( k \)-th position in the vector being 1 if node \( k \) is selected. In that case, $w_k(t)=i_k(t)=1$.


\subsubsection{Transition probability}
The transition between states relies on the 3 components of the state space and the components of the action space. The AoI is updated as follows \begin{equation}
\label{AOI_CALC}
	A_{k}(t+1) =
	\begin{cases}
		1, & \quad \text{if $i_k(t+1)=1$ }, \\
		A_{k}(t)+ 1, & \quad \text{otherwise}, 
	\end{cases}
\end{equation}
where \( i_k(t+1) \) is the element of the sparse vector $I(t+1)$.
The position of the AUV $l_{\text{auv}}(t)$ is updated according to the selected action $d(t)$ 
\begin{equation} \label{direction}
	l_{\text{auv}}(t+1)=
	\begin{cases}
		l_{\text{auv}}(t)+(d_g,0,0), & \quad d(t)=\text{right}, \\
		l_{\text{auv}}(t)-(d_g,0,0), & \quad d(t)=\text{left}, \\
		l_{\text{auv}}(t)+(0,d_g,0), & \quad d(t)=\text{up}, \\
		l_{\text{auv}}(t)-(0,d_g,0), & \quad d(t)=\text{down}, \\
  	l_{\text{auv}}(t)+(0,0,d_g), & \quad d(t)=\text{forward}, \\
		l_{\text{auv}}(t)-(0,0,d_g), & \quad d(t)=\text{backward}, \\
	\end{cases}
\end{equation} 
where $d_g$ is the unit distance between two squares in the grid.
As for the energy stored at each node, the update in a node $k$ is as follows
\begin{equation} \label{energy updating}
    \setlength{\arraycolsep}{2pt} 
    \boldsymbol{\mathcal{E}}_{k}(t+1) \! = \!
    \begin{cases}
        \boldsymbol{\mathcal{E}}_{k}(t) + e_{r}(t) - e_{c}(t), & \begin{aligned}[t] 
             \text{if } & w_k(t+1)=  \\
            &  i_k(t+1) = 1,
        \end{aligned} \\
        \boldsymbol{\mathcal{E}}_{k}(t) + e_{r}(t), & \text{if only } w_k(t+1) = 1, \\
        \boldsymbol{\mathcal{E}}_{k}(t) - e_{c}(t), & \text{if only } i_k(t+1) = 1, \\
        \boldsymbol{\mathcal{E}}_{k}(t), & \text{otherwise},
    \end{cases}
\end{equation}

where $e_{r}(t)$ is the energy received by the node from the AUV at time slot $t$ and $e_{c}(t)$ is the energy consumed for transmission of the data.
\subsubsection{Reward function} 
The reward system is defined to minimize the weighted sum of the AoI for all IoUT devices and to maximize Jain's fairness index. For mathematical convenience, we consider the discrimination index instead of the fairness that is defined in \eqref{jain fairness J}
\begin{equation}
\label{discrimination index}
\mathscr{D}_T = 1 - \mathcal{J}_T=1-\frac{\left(\sum_{k=1}^{K} D_T(k)\right)^2}{K \sum_{k=1}^{K} D_T(k)^2}.
\end{equation}
Herein, maximizing Jain's fairness index is equivalent to minimizing Jain's discrimination index.
We define the immediate reward $r_u$ for the AUV at time instant $t$ as 
\begin{equation}\label{reward function}
r_u(t) = -\left(\frac{\mathscr{D}_t}{K} \sum_{k=1}^{K} \delta_k A_k(t) +\rho(t) \right),
\end{equation}
with \(\rho(t)\) being the penalty term that varies based on the action space as follows
\begin{equation}
\label{penalty term}
\rho(t) = \rho_{\text{location}}(t) + \rho_{\text{information}}(t) + \rho_{\text{occurrence}}(t),
\end{equation}
where  \(\rho_{\text{location}}(t)\) is a penalty applied if the AUV is outside the designated set of valid locations \(\mathcal{X}\). It is defined as
\begin{equation}
\label{penalty for location}
\rho_{\text{location}}(t) = 
\begin{cases}
\rho_{\text{loc}}, & \text{if } l_{\text{auv}}(t) \notin \mathcal{X}, \\
0, & \text{otherwise}.
\end{cases}
\end{equation}
\(\rho_{\text{occurrence}}(t)\) is applied to avoid a scenario where a specific node \(k\) communicates with the AUV more frequently than others. The number of times node \(k\) has been selected to transmit information to the AUV by timeslot \(t\) is denoted by \(\mathcal{OCC}(k, t)\). The penalty is incurred if, at timeslot \(t\), the occurrence \(\mathcal{OCC}(k, t)\) is higher than the average communication time per node. Since \(T\) is the total navigation time of the AUV, \(\frac{T}{K}\) represents the ideal number of times each node would send data to the AUV if all nodes were selected equally. The penalty is defined as
\begin{equation}
\label{penalty for occurence}
\rho_{\text{occurrence}}(t) = 
\begin{cases}
\rho_{\text{occ}}, & \text{if } \mathcal{OCC}(k, t) > \frac{T}{K}, \\
0, & \text{otherwise}.
\end{cases}
\end{equation}
The penalty terms $\rho_{\text{location}}(t)$ and $\rho_{\text{occurrence}}(t)$ are the same for both FDD and TDD.  \\
\textbf{Penalty Term \(\rho_{\text{information}}(t)\) for 3D Action Space (FDD)}

This penalty is incurred when either no node is available to transmit data or the AUV makes an incorrect selection of a node for data collection. Let \(\mathcal{I}(t)\) denote the indices of the nodes that have sufficient energy to transmit data at timeslot \(t\). The penalty is defined as:
\begin{equation}
\label{penalty for missed information fdd}
\rho_{\text{information}}(t) = 
\begin{cases}
\rho_{\text{no\_indices}}, & \text{if } \mathcal{I}(t) = \varnothing, \\
\rho_{\text{wrong\_indice}}, & \text{if} \ k \notin \mathcal{I}(t) \text{ and } \mathcal{I}(t) \neq \varnothing, \\
0, & \text{otherwise}.
\end{cases}
\end{equation}
If no nodes have sufficient energy to transmit data (\(\mathcal{I}(t) = \varnothing\)), a penalty \(\rho_{\text{no\_indices}}\) is applied. If there exist nodes with sufficient energy (\(\mathcal{I}(t) \neq \varnothing\)), but the AUV selects a node that lacks energy for transmission, a penalty \(\rho_{\text{wrong\_indice}}\) is applied.
\textbf{Penalty Term \(\rho_{\text{information}}(t)\) for 2D Action Space (TDD)}

This penalty is introduced to account for situations where simultaneous AET and information uplink do not occur. That is, when the AUV is either only charging (i.e., IoUT node does not have enough energy to transmit data) or only collecting data (IoUT node has sufficient energy to transmit data without getting charged), rather than doing both. The penalty term is defined as
\begin{equation}
\label{penalty for missed information tdd}
\rho_{\text{information}}(t) = 
\begin{cases}
\rho_{\text{only\_charging}}, & \text{if } \beta = 1, \\
\rho_{\text{only\_transmitting}}, & \text{if } \beta = 0, \\
0, & \text{otherwise}.
\end{cases}
\end{equation}

\subsection{DRL Solution: Proximal Policy Optimization}
Proximal Policy Optimization (PPO) is a mix of two methods: a gradient-based policy that aims to maximize the reward through gradient descent and an actor-critic approach frequently utilized in RL. PPO stands out for its insensitivity to perturbations, a trait it addresses by constraining updates to the neural network. This is achieved by performing updates based on the ratio of the probability of the new policy to the old one. Additionally, PPO considers an advantage function to evaluate the value of each state \cite{schulman2017proximalpolicyoptimizationalgorithms}. By prioritizing profitable states while controlling the loss function, PPO employs techniques like clipping and setting a lower bound using a minimum function. One distinguishing feature of PPO is its approach to memory management. Instead of storing and sampling from millions of transitions randomly, PPO maintains a fixed-length trajectory of memories, simplifying the process.

The actor is responsible for the actions of the agent based on a learned policy that aims to minimize the clipped loss function defined as \cite{schulman2017proximalpolicyoptimizationalgorithms}
\begin{equation}
\label{loss function clipped}
L^{C L I P}(\theta)\!=\!\hat{\mathbb{E}}_{t}\left[\min \left(r_{t}(\theta) \hat{A}_{t}, \operatorname{clip}\left(r_{t}(\theta), 1\!-\!\epsilon, 1+\epsilon\right) \!\hat{A}_{t}\!\right)\right],
\end{equation}
where $r_{t}(\theta)$ denotes the probability ratio $r_{t}(\theta)=\frac{\pi_{\theta}\left(a_{t} \mid s_{t}\right)}{\pi_{\theta_{\text {old }}}\left(a_{t} \mid s_{t}\right)}$. Herein, $\pi_{\theta}(a_{t} | s_{t})$ is the new policy's probability of taking action $a_{t}$ in state $s_{t}$, and $\pi_{\theta_{\text{old}}}(a_{t} | s_{t})$ is the old policy's probability of taking action $a_{t}$ in state $s_{t}$. The term $\operatorname{clip}(r_{t}(\theta), 1 - \epsilon, 1 + \epsilon)$ is used to clip the ratio $r_{t}(\theta)$ between the lower and upper bounds of $1 - \epsilon$ and $1 + \epsilon$, respectively. The hyperparameter $\epsilon$ determines the extent of the clipping, and $\epsilon \approx 0.2$ is the most common value selected in the literature.
The objective is to maximize the clipped advantage function, which is a surrogate for the true advantage function \(\hat{A}_{t}\). The advantage function is defined as
\begin{multline}
\hat{A}_t = -V(s_t) + r_u(t) + \gamma r_u(t+1) + \cdots \\
+ \gamma^{T - t + 1} r_u(T-1) + \gamma^{T - t} V(s_T)
\end{multline}

where \(\gamma\) is the discount factor that determines the weight of future rewards, \(V(s_t)\) is the state value function at state \(s_t\), representing the expected return starting from state \(s_t\). \(r_u(t)\) is the reward received at time step \(t\), \(T\) is the final time step in the episode, and \(s_T\) is the state at the final time step \(T\). The Total loss is expressed as 
\begin{equation}
\label{total loss function}
L_{total}(\theta)=\hat{\mathbb{E}}_{t}\left[L_{t}^{C L I P}(\theta)-c_{1} L_{t}^{V F}(\theta)+c_{2} S\left[\pi_{\theta}\right]\left(s_{t}\right)\right] 
\end{equation}
where $c_{1}, c_{2}$ are coefficients used for exploitation and exploration respectively, and $S$ denotes an entropy bonus, and $L_{t}^{V F}$ is a squared-error loss $\left(V_{\theta}\left(s_{t}\right)-V_{t}^{\mathrm{targ}}\right)^{2}$. Algorithm 1 illustrates the PPO approach for both TDD and FDD\footnote{For simplicity, we present the two algorithms combined here. However, it is important to note that each algorithm is fundamentally different from the other. The key distinction lies in their action spaces, resulting in differing dimensions. Furthermore, each algorithm has its own set of penalty terms and requires separate hyperparameter tuning ( such as discount factor $\gamma$ and learning rate $\alpha$) to achieve optimal performance.}.
\begin{algorithm} [!t]
\caption{PPO for TDD and FDD }
\KwIn{Environment with locations of IoUT nodes $c_k$, AUV starting position $l_{auv}(0)$, AoI for all nodes $\boldsymbol{A}(0)$, other parameters (i.e, operating frequencies, etc..)}.
\KwOut{Trained PPO agent} 
\textbf{Initialization:} \\
Initialize PPO agent $\pi_\theta$, value network $V_\theta$\\
Set training parameters:  $max\_episodes$, $max\_iterations$..., \\

\textbf{Procedure of Training:}\\
\For{each episode $i = 1$ to $max\_episodes$}{
    $t \leftarrow 0$ \\
    Reset environment: $l_{auv}$, $\boldsymbol{A}(t)$, $\boldsymbol{\mathcal{E}}(t)$ and set $max\_iterations$  \\
    Initialize cumulative reward $R \leftarrow 0$  \\
    \While{$max\_iterations > 0$}{
        $t \leftarrow t + 1$ \\
                $max\_iterations \leftarrow max\_iterations - 1$ \\
        \textbf{Action Selection:}\\
        Observe state $s(t)$\\
        \eIf{TDD}{
            Sample action $a_t= \left\{d(t), I(t)=W(t)\right\}$\\
        }{
            Sample action $a_t= \left\{d(t), I(t),W(t)\right\}$\\
        }
        \textbf{Environment Interaction:}\\
        Move AUV according to the direction from $a_\tau$. \\
        Compute harvested energy and update AoI for \\ the  selected node(s) using \eqref{direction},\eqref{AOI_CALC},\eqref{energy updating}.\\
        
        \textbf{Reward Calculation:}\\
        Compute discrimination index using \eqref{discrimination index}. \\
        Compute reward $r_u(t)$ and penalize unfair AoI distribution and wasted energy according to the mode selected (FDD  or TDD) using equations \eqref{reward function}, \eqref{penalty for location}, \eqref{penalty for occurence}, \eqref{penalty for missed information fdd}, and \eqref{penalty for missed information tdd}.\\
        
        \textbf{Learning:}\\
        Store $(s_t, a_t, r_t, s_{t+1})$ in buffer \\
        \If{buffer full}{
            Update PPO policy $\pi_\theta$ and value function \\ $V_\theta$.
        }
        Update cumulative reward\\
        
    }
}
\textbf{Procedure of Testing:}\\
Load trained policy $\pi_\theta$ and evaluate on the test \\environment.\\
Simulate AUV operation and measure cumulative \\ reward,  fairness, AoI reduction.
\end{algorithm}

\section{Numerical Results}\label{sec:results}
We begin by outlining the assumptions and system configuration of our simulations, including the environment parameters and grid world setup. The problem is addressed using the PPO algorithm to evaluate the performance of both 3D-FDD and 2D-TDD approaches. We compare the results obtained from the proposed RL approaches with benchmark methods, such as RW, RR, and GA, to highlight the improvements and effectiveness of the proposed solutions.
\subsubsection{Simulation parameters}
In this study, we assumed a 3D grid for our simulations to better represent real-world underwater environments, which often involve a vertical dimension critical to communication and navigation. This choice adds complexity compared to the commonly used 2D grids, making the simulations more realistic. The grid size is set to \(1000 \, \text{m} \times 1000 \, \text{m} \times 400 \, \text{m}\), reflecting a reasonable approximation.It is discretized by a cell size of $100$ m, resulting in 400 unique geometric positions. The AUV begins at the center of this grid and communicates with K sensor nodes, where K $\in \{3, 5, 7, 10\}$.
 The main simulation parameters are depicted in Table \ref{tab:parameters}.
 \subsubsection{Algorithm architecture and hyperparameter settings}
For training, we employed the PPO algorithm inspired by the one implemented in the Stable Baselines3 library. The neural network architecture consists of two fully connected layers with 64 neurons each, using Tanh activation functions.  
To ensure consistency across varying network sizes (K $\in \{3, 5, 7, 10\}$), we tuned the hyperparameters using grid search. Our goal was to identify a single set of hyperparameters for both algorithms (3D-FDD and 2D-TDD). The selected hyperparameters are summarized in Table~\ref{tab:hyperparams}.

As illustrated in Algorithm 1, both techniques (TDD and FDD) share a similar RL environment in terms of the state space. However, they differ in the action space and reward function, as defined in equations \eqref{acion space tdd},\eqref{action space fdd},\eqref{penalty for missed information tdd} and \eqref{penalty for missed information fdd}. This distinction in the action space allows the agent to choose different actions (i.e., direction and node selection) depending on the duplexing technique adopted. The reward function is also adjusted accordingly to reflect the differences in energy transfer and AoI optimization.

\begin{table}[t!]
\caption{ The Simulation Parameters}
\centering
\begin{tabular}{|l|l|}
\hline
\textbf{\rule{0pt}{3ex} Parameter} & \textbf{\rule{0pt}{3ex} Value}  \\
\hline
Network size & $1000 \, \text{m} \times 1000 \, \text{m} \times 400 \, \text{m}$ \\
\hline
Electrical power ($P_{\text{elec}}$) & $2000 \, \text{Watts}$ \\
\hline
FDD Mode Frequency: & \begin{tabular}[c]{@{}l@{}}Inf transmission: $30 \, \text{kHz}$ \\ Energy harvesting: $40 \, \text{kHz}$\end{tabular} \\
\hline
TDD Mode Frequency: & $40 \, \text{kHz}$ \\
\hline
Elect-acoustic coefficient ($\eta$) & $0.5$ \\
\hline
Directivity index $(DI)$ & $20 \, \text{dB}$ \\
\hline
Sensitivity $(RVS)$ & $-150 \, \text{dB re V}/\mu \text{Pa}$ \\
\hline
Speed of AUV $(V)$ & $4 \, \text{m/s}$ \\
\hline
Spreading factor ($k_{s}$) & $1.5$ \\
\hline
Resistance $(R\textsubscript{p})$ & $125 \, \Omega$ \\
\hline
Packet size $(L\textsubscript{t})$ & $100 \, \text{bytes}$ \\
\hline
Bandwidth $(B)$& $3000 \, \text{Hz}$ \\
\hline
Total navigation time $(T)$ & $2500 \, \text{s}$ \\
\hline
Noise power $(NL)$ & $30 \, \text{dB}$ \\
\hline
Minimum Fairness ($J_{\text{Tmin}}$) & $0.85$ \\
\hline
Maximum Age ($A_{\text{max}}$) & $50$ \\
\hline
\end{tabular}
\label{tab:parameters}
\end{table}

\begin{table}[h!]
\centering
\caption{Hyperparameter settings for 2D-TDD and 3D-FDD}
\label{tab:hyperparams}
\begin{tabular}{|l|c|c|}
\hline
\textbf{Hyperparameter}      & \textbf{2D-TDD}       & \textbf{3D-FDD}       \\ \hline
Discount factor ($\gamma$)   & 0.93                  & 0.92                  \\ \hline
Learning rate   ($\alpha$)             & 0.0003                & 0.0005       \\ \hline
Entropy coefficient ($c_2)$        & 0.01                  & 0.01                  \\ \hline
Batch size                   & 100                   & 100                   \\ \hline
Steps per update ($n\_steps$)& 100                   & 100                   \\ \hline
\end{tabular}
\end{table}

\subsubsection{TDD mode configuration}
\begin{figure}[t!]
	\centering
	\includegraphics[width=0.95\columnwidth]{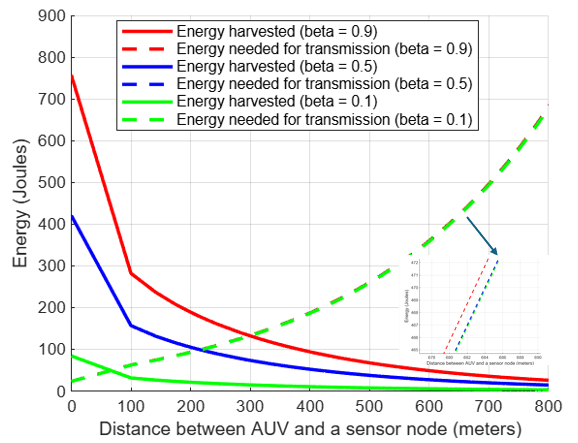}
	\caption{Energy harvested and information uplink for different Values of $\beta$ and for frequency $f=40$ KHz.}
	\label{Energies in TDD}
\end{figure}

In Fig.~\ref{Energies in TDD}, we analyze the TDD energy dynamics in the underwater communication system by examining the energy harvested and the energy required for information transmission across different values of \(\beta\) at a frequency of 40 KHz. The plot demonstrates that for \(\beta = 0.1\), the energy harvested is the lowest, as only a small portion of time is dedicated to energy harvesting. This configuration may lead to challenges in maintaining efficient long-distance transmission due to insufficient energy harvesting. As \(\beta\) increases to 0.5 and 0.9, the energy harvested significantly rises, highlighting the direct impact of \(\beta\) on energy allocation. In these scenarios, more time is dedicated to energy harvesting, allowing the AUV to sustain longer energy transmission effectively. Notably, the energy needed for transmission remains relatively unaffected by variations in \(\beta\). The observation that \(E_{\text{trans}}\) remains almost constant across different values of \(\beta\) is further explained in Appendix \ref{e_trans_constant}. 

\begin{figure}[!t]
	\centering
	\includegraphics[width=0.95\columnwidth]{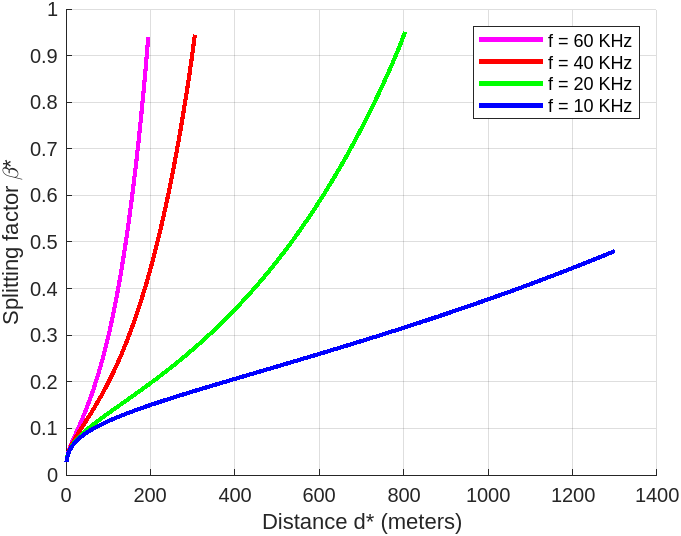}
    	\caption{Plot of $\beta^* $ as a function of $d^*$ for different frequencies}
        \label{beta for diff freq}
\end{figure}

Efficient simultaneous AET and information uplink operation is achieved when the energy harvested surpasses the energy required for transmission (i.e., the intersection points in Fig. \ref{Energies in TDD}). This critical condition is more likely to occur when \(\beta\) is higher, and the distance between the AUV and the IoUT node is short. To further explore the optimization of \(\beta\), Fig.~\ref{beta for diff freq} presents the relationship between \(\beta^*\) and the distance \(d^*\) for various communication frequencies which are the values of $\beta $ and $d$ that achieves the intersection in the Fig. \ref{Energies in TDD}. Here, we observe that \(\beta^*\) values ranging from 0 to 1 are achieved over shorter distances as the communication frequency increases. For instance, at higher frequencies such as 60 KHz, \(\beta^*\) quickly approaches its maximum value, reflecting a rapid signal attenuation requiring more charging time for reliable signal transmission.

Conversely, for lower frequencies, such as 10 KHz, \(\beta^*\) increases slowly with distance, supporting long-range AET and information uplink. This observation aligns with the goal of achieving a balanced configuration that supports both data transmission and energy harvesting, highlighting why a frequency of 40 KHz was selected in our next simulations. This medium frequency enables a practical compromise, optimizing the trajectory of the AUV by extending the range and reducing antenna size while maintaining adequate bandwidth for data and energy transfer.

Herein, we choose to restrict $\beta$ between 0.1 and 0.9 because practically, if the splitting factor is less than 0.1 (i.e., indicating that the charging time is less than 10\%), there will not be enough time for AET, and only information uplink occurs. Conversely, if the splitting factor exceeds 0.9 ($\beta > 0.9$), only charging occurs, with insufficient time for information uplink. In addition to the observations from Fig. \ref{beta for diff freq}, we provide a rigorous mathematical treatment of how $\beta$ is directly related to the distance $d$ through an inequality. This is required for simultaneous AET and information uplink to occur, ensuring that energy harvesting is sufficient for data transmission. The relationship between $\beta$ and $d$ is formalized in Appendix \ref{formula_for_f_d}, where we analyze the sufficient condition

\begin{equation}
   E_{harv} \geq E_{req,k},
\end{equation}
which leads to
\begin{equation}
    g(\beta) \geq h(d),
\end{equation}
where \(g(\beta)\) and \(h(d)\) are derived from energy harvesting and transmission models, as detailed in Appendix \ref{formula_for_f_d}.

By proving the function \(g(\beta)\) is bijective within the interval \(\beta_1 < \beta < \beta_2\), where $\beta_1$ is close to 0 and $\beta_2$ is close to 1 as shown in Appendix \ref{formula_for_f_d}, we demonstrate that \(\beta\) is strictly increasing with \(d\), implying a direct relationship. This bijection ensures that for any distance \(d^*\), there exists a unique \(\beta^*\), supporting our selection of frequencies and configurations in the simultaneous AET and information uplink process in TDD.

\subsubsection{Performance Evaluation}

\begin{figure}[t!]
	\centering
	\includegraphics[width=0.95\columnwidth]{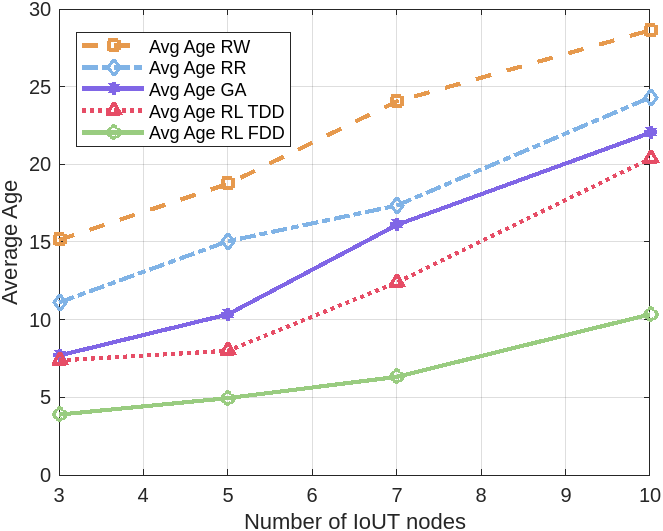}
	\caption{ Average AoI for different algorithms and network sizes.} 
	\label{Average AoI}
\end{figure}

The plot in Fig. \ref{Average AoI} illustrates the average AoI for various number of IoUT nodes using different AUV trajectory optimization and scheduling methods. The RW method, represented by the blue dashed line, shows the highest AoI across all nodes, indicating its inefficiency in maintaining up-to-date information. The RR method performs better than RW, reducing the AoI by approximately 16\% for small number of nodes and up to 25\% for a higher number of nodes (i.e, 7 and 10). The GA further reduces the AoI compared to RR, especially for lower-number of nodes with a reduction of approximately 30\%. This reflects that positioning the AUV near the center of gravity manages to solve the optimization problem more efficiently, but of course at the cost of fairness. In contrast, our RL based approaches, both 2D-TDD and 3D-TDD, demonstrate significant improvements over the GA and RR methods. The 2D-TDD achieves a lower AoI than GA, reducing it by about 15.5\%, which highlights its effectiveness in optimizing the AUV's path. The best performance is observed for the 3D-TDD, which maintains a consistently low AoI across all nodes to showcase its superior efficiency in terms of AoI reduction over all other methods. 

\begin{figure}[t!]
	\centering
	\includegraphics[width=0.95\columnwidth]{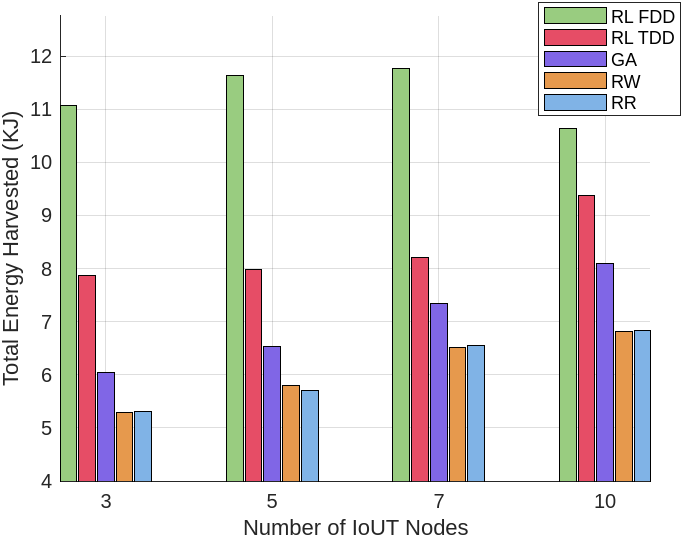}
	\caption{ Total energy harvested plot for different algorithms and network sizes. }
	\label{Average Energy Harvested}
\end{figure}

\begin{figure}[t!]
	\centering
	\includegraphics[width=0.95\columnwidth]{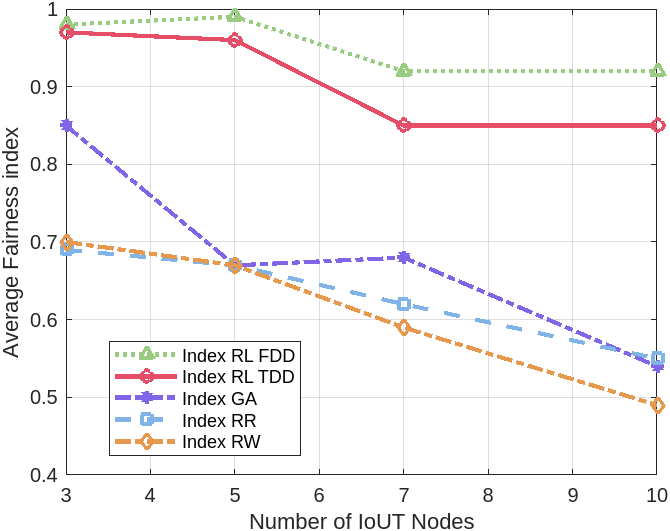}
	\caption{ The average Jain's fairness index for different algorithms and network sizes } 
	\label{Average Jain index}
\end{figure}

Looking at Fig. \ref{Average Energy Harvested}, we notice that both RW and RR methods render similar and relatively low energy harvesting capabilities across all node counts (i.e, not exceeding 7 KJ). This indicates their inefficiency in energy management. The proposed RL approaches with 2D-TDD and 3D-FDD action spaces demonstrate significant improvements over the RW, RR, and GA methods. The RL with a 2D-TDD action space harvests an amount ranging from 8 KJ to 9.2 KJ.  However, its performance is inferior to the 3D-TDD (green bars), which achieves the highest energy harvesting levels among all methods across all different network sizes, indicating its superiority as expected. The RL 3D-FDD method harvests more than 10.5 KJ in all cases and up to 11.7 KJ in the 7-node setup.  This highlights the 3D-FDD approach's effectiveness and superiority in energy harvesting by dynamically adjusting to the environment. We can further explore the efficiency of 3D-FDD in Fig. \ref{Average Jain index}, where it consistently maintains the highest fairness index outperforming 2D-TDD, which leads to a quasi-uniform distribution of resources across all nodes. In contrast, the benchmark methods exhibit increased randomness for a higher number of IoUT devices, resulting in lower fairness as shown in the figure.

\subsubsection{Complexity Evaluation} The environmental impact of training ML models has become a significant concern within the ML community due to their substantial energy consumption and carbon emissions.  The eco2ai library introduced in \cite{budennyy2022eco2ai} addresses this issue by providing a tool to track AI models' carbon footprint and energy consumption during both the training and inference phases, promoting sustainable practices in AI development.  In our work, we adopt eco2ai to compare the 2D-TDD to 3D-FDD. Table. \ref{Carbon Emission} highlights that models with more IoUT nodes generally require longer training times, leading to higher energy consumption and more CO2 emissions. The consistent use of the same hardware (i.e, CPU: AMD Ryzen 7 5800H with  Radeon Graphics; GPU: NVIDIA GeForce GTX 1650 ) across all models ensures that these variations are primarily due to differences in model configurations and training times. Notably, the model that adopts 2D-TDD consistently consumes less energy and emits less CO2 compared to their 3D-FDD counterparts, indicating a lower environmental impact for simpler models.

\begin{table}[t!]
\caption{ A Complexity Analysis and Environmental Impact for the Proposed Methods.}
\centering
\begin{tabular}{|l|l|l|l|}
\hline
\textbf{\rule{0pt}{3ex} PPO Model} & \textbf{\rule{0pt}{3ex} Train Time} & \textbf{\rule{0pt}{3ex} Energy (kWh)} & \textbf{\rule{0pt}{3ex} CO2 (kg)} \\ \hline
3 \text{ nodes TDD} & 58 \text{ min} & 0.0173 & 0.0099 \\ \hline
3 \text{ nodes FDD} & 1 \text{ h 9 min} & 0.0188 & 0.0107 \\ \hline
5 \text{ nodes TDD} & 1 \text{ h 24 min} & 0.0231 & 0.0132 \\ \hline
5 \text{ nodes FDD} & 2 \text{ h 17 min} & 0.0838 & 0.0478 \\ \hline
7 \text{ nodes TDD} & 3 \text{ h 34 min} & 0.1382 & 0.0788 \\ \hline
7 \text{ nodes FDD} & 6 \text{ h 4 min} & 0.2174 & 0.1240 \\ \hline
10 \text{ nodes TDD} & 6 \text{ h 47 min} & 0.2246 & 0.1260 \\ \hline
10 \text{ nodes FDD} & 8 \text{ h 11 min} & 0.2942 & 0.1679 \\ \hline
\end{tabular}

\label{Carbon Emission}

\end{table}

While the current study is based on simulations, the proposed solutions can be tested in real-world scenarios to validate their applicability further. Specifically, the framework can be implemented in a controlled underwater testbed using AUVs equipped with acoustic modems for communication and energy transfer modules. Such an experimental setup would allow us to evaluate the practical feasibility of the FDD and TDD schemes under varying underwater conditions, including different water depths, noise levels, and environmental factors.
\section{Conclusions} \label{sec:conclusions}
This paper proposed two RL schemes for simultaneous AET and information uplink in IoUT network via AUV. The first scheme is FDD, which shows high performance in terms of age minimization and energy harvesting at the cost of higher complexity and hardware cost. The second scheme is TDD which offers a sub-optimal performance at lower complexity, less Co2 emissions and hardware cost. The results demonstrated that both schemes significantly improved energy harvesting efficiency, data collection, and fairness in resource distribution compared to conventional RW, GA, RR benchmarks. 

Future work may include accounting for the AUV navigation energy and the deployment of multiple AUVs to serve massive IoUT networks via multi-agent reinforcement learning and meta-learning for dynamic environments. Reconfigurable intelligent surfaces (RIS) could be also applied in underwater environments to boost the efficiency of AET and information uplink in the same way it is deployed in terrestrial and non-terrestrial networks \cite{swipt_plus}.

\appendices 
\section{ Demonstrating how \( E_{req,k} \) is almost constant}
\label{e_trans_constant}
    In this appendix, we provide a detailed explanation for why the energy needed for transmission, \(E_{req,k}\), remains relatively unaffected by changes in \(\beta\). We start with the expression for \(E_{req,k}\)
\begin{equation}
E_{req,k}= P_{trans,k}\cdot (1-\beta)\cdot \tau.
\end{equation} 
Using \eqref{p transmit}
\begin{equation}
E_{req,k} = \gamma_{\mathrm{k}} \cdot 10^{\frac{AL + NL}{10}} \cdot (1-\beta)\tau.
\end{equation}
Referring to \eqref{snr}, we can write the throghput as
\begin{equation}
\mathcal{S} =  \frac{L_t}{(1-\beta) \cdot \tau},
\end{equation}
where \( L_t \) is the packet size, indicating the quantity of data per transmission

Hence,
\begin{equation}
\label{E_req_k_appendix}
E_{req,k} = \left(2^{\left(\frac{L_t}{(1-\beta) \cdot \tau \cdot B}\right)} - 1\right) \cdot 10^{\frac{AL + NL}{10}} \cdot (1-\beta)  \tau.
\end{equation}
Using the Taylor expansion, we can deduce that
\begin{equation}
2^{\left(\frac{L_t}{(1-\beta) \cdot \tau \cdot B}\right)} \approx 1 + \frac{L_t \cdot \log(2)}{(1-\beta) \tau B}.
\end{equation}
Therefore,
\begin{equation}
E_{req,k} \approx \frac{L_t \cdot \log 2}{(1-\beta) \cdot \tau \cdot B} \cdot 10^{\frac{AL + NL}{10}} \cdot (1-\beta) \tau
\end{equation}
Simplifying, we obtain 
\begin{equation}
E_{req,k} (\beta, d) \approx E_{req,k} (d) = \frac{L_t \cdot \log2}{B} \times 10^{\frac{AL(d) + NL}{10}}.
\end{equation}
Now, let $\text{\LARGE $\kappa$}=\frac{L_t \cdot \log(2)}{B}\cdot 10^{\frac{ NL}{10}}$. We attain
\begin{equation}
E_{req,k} (d) \approx \text{\LARGE $\kappa$} \cdot10^{\frac{AL(d)}{10}}.
\end{equation}
Plugging the values indicated in TABLE II, we end up with 
\begin{equation}
\label{e_req_approximation}
E_{req,k} (d) \approx 23.105 \times 10^{\frac{AL(d)}{10}}.
\end{equation}

Thus, the energy required for transmission is approximately constant due to the significant effect of the constant term in the power calculation and the nearly linear dependency of the SNR on \( (1 - \beta) \). From Table \ref{table:energy_comparison}, we can observe the tightness of the proposed approximation, where the normalized mean square error is in the order of $10^{-5}$ for all values of $\beta$ and $d$.
\begin{table}[ht]
\caption{A Comparison of True and Approximated Energy Values.}
\centering
\renewcommand{\arraystretch}{1.2} 
\begin{tabular}{|c|c|c|c|c|}
\hline
\textbf{Distance (d)} & \textbf{Approximated} & \multicolumn{3}{c|}{\textbf{True Energy (J)} \eqref{E_req_k_appendix} } \\

 & \textbf{Energy (J)  \eqref{e_req_approximation}} &  \(\beta = 0.1\) & \(\beta = 0.5\) & \(\beta = 0.9\) \\
\hline
100  & 62.0503   & 62.0819   & 62.1075   & 62.3377   \\
\hline
200  & 92.6695   & 92.7167   & 92.7548   & 93.0986   \\
\hline
300  & 132.5527  & 132.6202  & 132.6747  & 133.1665  \\
\hline
700  & 494.0117  & 494.2635  & 494.4666  & 496.2996  \\
\hline
800  & 678.3817  & 678.7274  & 679.0063  & 681.5234  \\
\hline
900  & 929.3624  & 929.8360  & 930.2181  & 933.6665  \\
\hline
\end{tabular}
\label{table:energy_comparison}
\end{table}

\section{Energy Harvesting vs energy required for transmission }

\label{formula_for_f_d}
We explore the relationship between \(\beta\) and \(d\), ensuring that energy harvesting is sufficient for reliable data transmission.
\subsubsection{Energy harvested} Assuming $\beta$ amount of time for energy harvesting and using \eqref{RL} in \eqref{p harvested}, we obtain 
\begin{equation}
\begin{split}
E_{harv} &= \beta \cdot \tau \cdot P_{harv} \\
&= \beta \cdot \tau \cdot  \frac{\eta}{4R_p} 10^{\frac{RVS + RL}{10}} \\
&= \beta \cdot \tau \cdot \text{\large $\kappa_2$ } \cdot 10^{-\frac{AL}{10}},
\end{split}
\end{equation}
 \text{where}  \( \kappa_2 \) $=  \frac{\eta}{4R_p} 10^{\frac{RVS + SL - NL}{10}}$.

\subsubsection{Energy required for transmission} Assuming $1-\beta$ portion of time for information uplink, from  \eqref{E_req_k_appendix} we have 
\begin{equation} \label{ereq}
E_{req,k} = \left(2^{\left(\frac{L_t}{(1-\beta) \cdot \tau \cdot B}\right)} - 1\right) \times 10^{\frac{AL + NL}{10}} \times (1-\beta) \cdot \tau.
\end{equation}

\subsubsection{Inequality Comparison} In order to perform successful transmission, the harvested energy must be greater than or equal to the amount of energy required for the information uplink. That is 
\begin{align} \label{eharv}
E_{harv} &\geq E_{req,k},
\end{align}
Plugging \eqref{ereq} and \eqref{eharv}, the inequality gives
\begin{equation}
\beta \cdot \kappa_2 \cdot 10^{-\frac{AL}{10}} \geq (1 - \beta) \cdot \left(2^{\smash{\scriptstyle \frac{L_t}{(1-\beta) \cdot \tau \cdot B}}} - 1\right) \cdot 10^{\smash{\scriptstyle \frac{AL + NL}{10}}}
\end{equation}
leading to
\begin{equation}
\frac{\beta \cdot \kappa_3}{(1 - \beta) \left(2^{\smash{\scriptstyle \frac{L_t}{(1-\beta) \cdot \tau \cdot B}}} - 1\right)} \geq 10^{\smash{\scriptstyle \frac{AL}{5}}}, \quad \kappa_3 = \kappa_2 \cdot 10^{\smash{\scriptstyle -\frac{NL}{10}}}
\end{equation}

Taking logarithm  base 10 for both sides, we attain
\begin{equation}
\begin{aligned}
\log_{10} \left( \frac{\beta}{1 - \beta} \right) - \log_{10} \left( 2^{\smash{\scriptstyle \frac{L_t}{(1-\beta) \cdot \tau \cdot B}}} - 1 \right) &\geq \frac{AL}{5} - \log_{10} \kappa_3,
\end{aligned}
\end{equation}
where 
\begin{align}
AL(d) &= k_s \cdot\log_{10}(d) +  \kappa_4 \cdot d  \quad \text{with} \quad \kappa_4 = \alpha(f).
\end{align}

Finally, 
\begin{equation}
\begin{aligned}
\log_{10} \left( \frac{\beta}{1 - \beta} \right) - \log_{10} &\left( 2^{\smash{\scriptstyle \frac{L_t}{(1-\beta) \cdot \tau \cdot B}}} - 1 \right)  \geq \\
&  \frac{k_s \log_{10}(d)}{5} + \frac{\kappa_4 d}{5} - \log_{10} \kappa_3.
\end{aligned}
\end{equation}

Hence,
\begin{equation}
g(\beta) \geq h(d),
\end{equation}
\text{where}
\begin{equation}
g(\beta) = \log_{10}\left(\frac{\beta}{1 - \beta}\right) - \log_{10}\left(2^{\frac{L_t}{(1 - \beta) \cdot \tau \cdot B}} - 1\right)
\end{equation}
\begin{equation}
h(d) = \frac{1}{5} \left( k_s \log_{10}(d) + \kappa_4 \cdot d   \right) - \log_{10} \kappa_3
\end{equation}

\subsubsection{Derivative Calculation}



The derivation of $g(\beta)$ with respect to $\beta$ gives
\begin{equation}
\begin{aligned}
\frac{\partial g}{\partial \beta} = \frac{1}{\log10}& \left( \frac{1}{\beta(1 - \beta)} \right. \\
&\quad + \left. \frac{2^{\smash{\scriptstyle \frac{L_t}{(1 - \beta) \cdot \tau \cdot B}}} \log2 \cdot \left(\frac{L_t}{(1 - \beta)^2 \cdot \tau \cdot B}\right)}{2^{\smash{\scriptstyle \frac{L_t}{(1 - \beta) \cdot \tau \cdot B}}} - 1} \right).
\end{aligned}
\end{equation}

\subsubsection*{Proof of Bijectivity}
To prove the bijectivity of \( g(\beta) \), we need to prove its injectivity and surjectivity within the domain \( \beta \in [\beta_1, \beta_2] \) where \( \beta_1 \) is close to 0 and \( \beta_2 \) is close to 1. To prove injectivity, we examine the derivative of \( g(\beta) \). Since \( \beta \in [\beta_1, \beta_2] \), both \( \beta \) and \( 1 - \beta \) are positive, making \( \frac{1}{\beta(1 - \beta)} > 0 \). For the second term, the expression \( \frac{L_t}{(1 - \beta) \cdot \tau \cdot B} > 0 \) since \( L_t, \tau, \) and \( B \) are all positive. Therefore, \( 2^{\frac{L_t}{(1 - \beta) \cdot \tau \cdot B}} > 1 \), which implies that both the numerator \( 2^{\frac{L_t}{(1 - \beta) \cdot \tau \cdot B}} \cdot \log(2) \cdot \left(\frac{L_t}{(1 - \beta)^2 \cdot \tau \cdot B}\right) \) and the denominator \( 2^{\frac{L_t}{(1 - \beta) \cdot \tau \cdot B}} - 1 \) are positive. Hence, the overall derivative \( \frac{\partial g}{\partial \beta} > 0 \) in the interval \( \beta \in [\beta_1, \beta_2] \), confirming that \( g \) is strictly increasing and injective in the interval \( \beta \in [\beta_1, \beta_2] \).

\subsubsection*{Proof of Surjectivity}
Since \( g(\beta) \) is continuous by definition of logarithmic function, and strictly increasing in the interval \( \beta \in [\beta_1, \beta_2] \), it will cover all values between its minimum and maximum, \( g(\beta_1) \) and \( g(\beta_2) \). Therefore, \( g \) is surjective onto its range.

Since \( g \) is both injective and surjective in the interval \( \beta \in [\beta_1, \beta_2] \), we conclude that \( g(\beta) \) is bijective in this interval.


%
\bibliographystyle{IEEEtran}
\bibliography{references}

@ARTICLE{10268591,
  author={Omeke, Kenechi G. and Mollel, Michael and Shah, Syed T. and Zhang, Lei and Abbasi, Qammer H. and Imran, Muhammad Ali},
  journal={IEEE Internet of Things Journal}, 
  title={Toward a Sustainable Internet of Underwater Things Based on AUVs, SWIPT, and Reinforcement Learning}, 
  year={2024},
  volume={11},
  number={5},
  pages={7640-7651},
  doi={10.1109/JIOT.2023.3319250}}

@misc{schulman2017proximalpolicyoptimizationalgorithms,
      title={Proximal Policy Optimization Algorithms}, 
      author={John Schulman and Filip Wolski and Prafulla Dhariwal and Alec Radford and Oleg Klimov},
      year={2017},
      eprint={1707.06347},
      archivePrefix={arXiv},
      primaryClass={cs.LG},
      url={https://arxiv.org/abs/1707.06347}, 
}

@ARTICLE{6266681,
  author={Bereketli, Alper and Bilgen, Semih},
  journal={IEEE Sensors Journal}, 
  title={Remotely Powered Underwater Acoustic Sensor Networks}, 
  year={2012},
  volume={12},
  number={12},
  pages={3467-3472},
  doi={10.1109/JSEN.2012.2210401}}

@ARTICLE{9217956,
  author={Guida, Raffaele and Demirors, Emrecan and Dave, Neil and Melodia, Tommaso},
  journal={IEEE Transactions on Mobile Computing}, 
  title={Underwater Ultrasonic Wireless Power Transfer: A Battery-Less Platform for the Internet of Underwater Things}, 
  year={2022},
  volume={21},
  number={5},
  pages={1861-1873},
  doi={10.1109/TMC.2020.3029679}}

@article{centelles2015wireless,
  title={{Wireless RF camera monitoring for underwater cooperative robotic archaeological applications}},
  author={Centelles, Diego and Rubino, Eduardo M and Sales, Jorge and Mart{\'\i}, Jos{\'e} V and Mar{\'\i}n, Ra{\'u}l and Sanz, Pedro J},
  journal={Instrumentation viewpoint},
  number={18},
  pages={51--52},
  year={2015},
  publisher={SARTI}
}

@article{pal2022communication,
  title={Communication for underwater sensor networks: A comprehensive summary},
  author={Pal, Amitangshu and Campagnaro, Filippo and Ashraf, Khadija and Rahman, Md Rashed and Ashok, Ashwin and Guo, Hongzhi},
  journal={ACM Transactions on Sensor Networks},
  volume={19},
  number={1},
  pages={1--44},
  year={2022},
  publisher={ACM New York, NY}
}

@article{shihada2020aqua,
  title={Aqua-Fi: Delivering Internet underwater using wireless optical networks},
  author={Shihada, Basem and Amin, Osama and Bainbridge, Christopher and Jardak, Seifallah and Alkhazragi, Omar and Ng, Tien Khee and Ooi, Boon and Berumen, Michael and Alouini, Mohamed-Slim},
  journal={IEEE Communications Magazine},
  volume={58},
  number={5},
  pages={84--89},
  year={2020},
  publisher={IEEE}
}

@misc{ahmadian,
      title={{Wireless Powered Communication Networks: TDD or FDD?}}, 
      author={Arman Ahmadian and Hyuncheol Park},
      year={2018},
      eprint={1807.05670},
      archivePrefix={arXiv},
      primaryClass={cs.IT},
      url={https://arxiv.org/abs/1807.05670}, 
}

@ARTICLE{9950310,
  author={Eldeeb, Eslam and Sant'Ana, Jean Michel de Souza and Perez, Dian Echevarría and Shehab, Mohammad and Mahmood, Nurul Huda and Alves, Hirley},
  journal={IEEE Transactions on Vehicular Technology}, 
  title={Multi-UAV Path Learning for Age and Power Optimization in IoT With UAV Battery Recharge}, 
  year={2023},
  volume={72},
  number={4},
  pages={5356-5360},
  doi={10.1109/TVT.2022.3222092}}

@ARTICLE{extreme,
  author={Guo, Hongzhi and Ofori, Albert A.},
  journal={IEEE Internet of Things Magazine}, 
  title={The Internet of Things in Extreme Environments Using Low-Power Long-Range Near Field Communication}, 
  year={2021},
  volume={4},
  number={1},
  pages={34-38},
  doi={10.1109/IOTM.0011.2000063}}

@inproceedings{budennyy2022eco2ai,
  title={{Eco2AI: carbon emissions tracking of machine learning models as the first step towards sustainable AI}},
  author={Budennyy, Semen Andreevich and Lazarev, Vladimir Dmitrievich and Zakharenko, Nikita Nikolaevich and Korovin, Aleksei N and Plosskaya, OA and Dimitrov, Denis Valer’evich and Akhripkin, VS and Pavlov, IV and Oseledets, Ivan Valer'evich and Barsola, Ivan Segundovich and others},
  booktitle={Doklady Mathematics},
  volume={106},
  number={Suppl 1},
  pages={S118--S128},
  year={2022},
  organization={Springer}
}

@article{AoI,
  author    = {Antzela Kosta and Nikolaos Pappas and Vangelis Angelakis},
  title     = {Age of information: A new concept, metric, and tool},
  journal   = {Foundations and Trends in Networking, Now Publishers, Inc.},
  year      = {2017},
}

@INPROCEEDINGS{Magid,
  author={Abd-Elmagid, Mohamed A. and Ferdowsi, Aidin and Dhillon, Harpreet S. and Saad, Walid},
  booktitle={2019 IEEE Global Communications Conference (GLOBECOM)}, 
  title={{Deep Reinforcement Learning for Minimizing Age-of-Information in UAV-Assisted Networks}}, 
  year={2019},
  volume={},
  number={},
  pages={1-6},
  doi={10.1109/GLOBECOM38437.2019.9013924}}

@ARTICLE{Jouhari,
  author={Jouhari, Mohammed and Ibrahimi, Khalil and Tembine, Hamidou and Ben-Othman, Jalel},
  journal={IEEE Access}, 
  title={Underwater Wireless Sensor Networks: A Survey on Enabling Technologies, Localization Protocols, and Internet of Underwater Things}, 
  year={2019},
  volume={7},
  number={},
  pages={96879-96899},
  doi={10.1109/ACCESS.2019.2928876}}

@ARTICLE{10012479,
  author={Omeke, Kenechi G. and Abubakar, Attai I. and Zhang, Lei and Abbasi, Qammer H. and Imran, Muhammad Ali},
  journal={IEEE Internet of Things Magazine}, 
  title={How Reinforcement Learning is Helping to Solve Internet-of-Underwater-Things Problems}, 
  year={2022},
  volume={5},
  number={4},
  pages={24-29},
  doi={10.1109/IOTM.001.2200129}}

@article{article,
author = {Reddy, Srinivasa and Arya, Rajeev and Prateek},
year = {2024},
month = {08},
pages = {1-1},
title = {Compensation of Coordinated Attacks in Underwater Internet of Sensor Networks},
volume = {PP},
journal = {IEEE Transactions on Consumer Electronics},
doi = {10.1109/TCE.2024.3432785}
}

@Inbook{Safeer2024,
author="Safeer, Ehtesham
and Tahir, Sidra
and Shaheen, Momina
and Farooq, Muhammad Shoaib",
editor="De, Debashis
and Sengupta, Diganta
and Tran, Tien Anh",
title="Federated Learning for Internet of Underwater Drone Things",
bookTitle="Artificial Intelligence and Edge Computing for Sustainable Ocean Health",
year="2024",
publisher="Springer Nature Switzerland",
address="Cham",
pages="295--309",
isbn="978-3-031-64642-3",
doi="10.1007/978-3-031-64642-3_13",
url="https://doi.org/10.1007/978-3-031-64642-3_13"
}

@article{loc2,
title = {Hierarchical localization algorithm for sustainable ocean health in large-scale underwater wireless sensor networks},
journal = {Sustainable Computing: Informatics and Systems},
volume = {39},
pages = {100902},
year = {2023},
issn = {2210-5379},
doi = {https://doi.org/10.1016/j.suscom.2023.100902},
url = {https://www.sciencedirect.com/science/article/pii/S2210537923000574},
author = {Tanveer Ahmad and Xue Jun Li and Aswani Kumar Cherukuri and Ki-Il Kim},
}

@INPROCEEDINGS{ultrasonic,
  author={Zhao, Yufei and Du, Yuwei and Wang, Zhenxing and Wang, Jianhua and Geng, Yingsan},
  booktitle={2021 IEEE Wireless Power Transfer Conference (WPTC)}, 
  title={Design of Ultrasonic Transducer Structure for Underwater Wireless Power Transfer System}, 
  year={2021},
  volume={},
  number={},
  pages={1-4},
  doi={10.1109/WPTC51349.2021.9458061}}

@ARTICLE{Survey,
  author={Wibisono, Arif and Alsharif, Mohammed H. and Song, Hyoung-Kyu and Lee, Byung Moo},
  journal={IEEE Access}, 
  title={A Survey on Underwater Wireless Power and Data Transfer System}, 
  year={2024},
  volume={12},
  number={},
  pages={34942-34957},
  doi={10.1109/ACCESS.2024.3373791}}

@ARTICLE{Relay,
  author={Dai, Jun and Li, Xinbin and Han, Song and Yu, Junzhi and Liu, Zhixin},
  journal={IEEE Transactions on Green Communications and Networking}, 
  title={Relay Selection and Power Control for Mobile Underwater Acoustic Communication Networks: A Dual-Thread Reinforcement Learning Approach}, 
  year={2024},
  volume={},
  number={},
  pages={1-1},
  doi={10.1109/TGCN.2024.3445142}}

@INPROCEEDINGS{swarm,
  author={Eldeeb, Eslam and Shehab, Mohammad and Alves, Hirley},
  booktitle={2023 IEEE 34th Annual International Symposium on Personal, Indoor and Mobile Radio Communications (PIMRC)}, 
  title={Age Minimization in Massive IoT via UAV Swarm: A Multi-agent Reinforcement Learning Approach}, 
  year={2023},
  volume={},
  number={},
  pages={1-6},
  doi={10.1109/PIMRC56721.2023.10293964}}

@article{zia2021state,
  title={State-of-the-art underwater acoustic communication modems: Classifications, analyses and design challenges},
  author={Zia, Muhammad Yousuf Irfan and Poncela, Javier and Otero, Pablo},
  journal={Wireless personal communications},
  volume={116},
  pages={1325--1360},
  year={2021},
  publisher={Springer}
}

@inproceedings{eid2023enabling,
  title={Enabling long-range underwater backscatter via van atta acoustic networks},
  author={Eid, Aline and Rademacher, Jack and Akbar, Waleed and Wang, Purui and Allam, Ahmed and Adib, Fadel},
  booktitle={Proceedings of the ACM SIGCOMM 2023 Conference},
  pages={1--19},
  year={2023}
}

@INPROCEEDINGS{9692982,
  author={Farag, Hossam and Gidlund, Mikael and Stefanovic, Cedomir},
  booktitle={2021 IEEE Global Conference on Artificial Intelligence and Internet of Things (GCAIoT)}, 
  title={A Deep Reinforcement Learning Approach for Improving Age of Information in Mission-Critical IoT}, 
  year={2021},
  volume={},
  number={},
  pages={14-18},
  doi={10.1109/GCAIoT53516.2021.9692982}}

@ARTICLE{AoI_plus,
  author={Ma, Yue and Ma, Ruiqian and Lin, Zhi and Zhang, Ruoyu and Cai, Yueming and Wu, Wen and Wang, Jiangzhou},
  journal={IEEE Internet of Things Journal (early access)}, 
  title={Improving Age of Information for Covert Communication With Time-Modulated Arrays}, 
  year={2024},
  volume={},
  number={},
  pages={1-1},
  doi={10.1109/JIOT.2024.3466855}}

@ARTICLE{swipt_plus,
  author={An, Kang and Sun, Yifu and Lin, Zhi and Zhu, Yonggang and Ni, Wanli and Al-Dhahir, Naofal and Wong, Kai-Kit and Niyato, Dusit},
  journal={IEEE Transactions on Wireless Communications}, 
  title={Exploiting Multi-Layer Refracting RIS-Assisted Receiver for HAP-SWIPT Networks}, 
  year={2024},
  volume={23},
  number={10},
  pages={12638-12657},
  doi={10.1109/TWC.2024.3394214}}

@ARTICLE{10032494,
  author={He, Yejun and Gan, Youhui and Cui, Haixia and Guizani, Mohsen},
  journal={IEEE Internet of Things Journal}, 
  title={Fairness-Based 3-D Multi-UAV Trajectory Optimization in Multi-UAV-Assisted MEC System}, 
  year={2023},
  volume={10},
  number={13},
  pages={11383-11395},
  doi={10.1109/JIOT.2023.3241087}}
\end{document}